# Scaling Artificial Intelligence for Digital Wargaming in Support of Decision-Making

**Scotty Black**
Naval Postgraduate School
1 University Circle, Monterey, CA 93943
UNITED STATES

scotty.black@nps.edu

**Christian Darken**
Naval Postgraduate School
1 University Circle, Monterey, CA 93943
UNITED STATES

cjdarken@nps.edu

## ABSTRACT

*In this unprecedented era of technology-driven transformation, it becomes more critical than ever that we aggressively invest in developing robust artificial intelligence (AI) for wargaming in support of decision-making. By advancing AI-enabled systems and pairing these with human judgment, we will be able to enhance all-domain awareness, improve the speed and quality of our decision cycles, offer recommendations for novel courses of action, and more rapidly counter our adversary's actions. It therefore becomes imperative that we accelerate the development of AI to help us better address the complexity of modern challenges and dilemmas that currently requires human intelligence and, if possible, attempt to surpass human intelligence—not to replace humans, but to augment and better inform human decision-making at machine speed. Although deep reinforcement learning continues to show promising results in intelligent agent behavior development for the long-horizon, complex tasks typically found in combat modeling and simulation, further research is needed to enable the scaling of AI to deal with these intricate and expansive state-spaces characteristic of wargaming for either concept development, education, or analysis. To help address this challenge, in our research, we are developing and implementing a hierarchical reinforcement learning framework that includes a multi-model approach and dimension-invariant observation abstractions.*

## 1.0 WARGAMING

Although a shared common definition of wargaming does not exist throughout the wargaming community, *Joint Publication 5-0 Joint Planning* [1] defines wargames as "representations of conflict or competition in a synthetic environment, in which people make decisions and respond to the consequences of those decisions." Other definitions include McHugh's definition in *U.S. Navy Fundamentals of Wargaming* [2], "a simulation, in accordance with predetermined rules, data, and procedures, of selected aspects of a conflict situation." In a RAND study [3], Wong et al. stated that a "wargame involves human players or actors making decisions in an artificial contest environment and then living with the consequences of their actions." We believe Caffrey [4] said it best when he wrote "[j]ust as in the old story of the blind men describing an elephant, definitions of wargaming tend to reflect who is doing the defining." Nevertheless, we think Perla's definition of *wargame* in *The Art of Wargaming* summarizes most definitions in a broad, yet nuanced, manner with "a warfare model or simulation whose operation does not involve the activities of actual military forces, and whose sequence of events affects and is, in turn, affected by the decisions made by the players representing the opposing sides" [5]. Additionally, Perla [6] defines *wargaming* (as opposed to *wargame*) as "[a]n applied discipline encompassing the creation, use, synthesis and analysis of wargames to conduct research, explore concepts, develop and test hypotheses, and dynamically communicate insights to inform, educate, and train individuals and organizations."

With all of these different definitions and concepts of a *wargame*, as well as the broad use cases for *wargaming*, it is easy to see that the practice of wargaming can encompass such disparate activities ranging from a table-top game on a physical board with pieces that represent units or weapons systems, to completely simulated operations within a complex virtual environment in the style of a video game [7].

While wargames are neither predictive nor complete replications of reality [8], they do offer something that cannot be obtained without actual combat: *insight into decision-making in war*. Wargaming enables players to engage with challenges to a degree that is difficult for non-wargamers to experience, and that produces substantive long-term gains in educational value [9]. Variation in combat outcomes during the game creates unpredictable threats and opportunities, which force players to make decisions on classic, real-world dilemmas [9]. They offer us a means to prepare decision makers for the complex and uncertain environments that we expect to encounter given the pace and depth of change in our current global dynamics [10]. Moreover, as Perla and McGrady point out in *Why Wargaming Works* [10], to get the most value out of wargames, we must integrate them with other tools—such as analysis, exercises, history, and real-world experiences—to help us better make sense of what we should do, whether in the present or in the future. Ultimately, like in real combat, "wargames can help us learn important things about uncertainty" [6], as well as help us better understand what we really know (rather than what we think we know), what we don't know, what we don't know we know, and—perhaps even—what we don't know we don't know [6].

## 1.1 A Call for Wargaming Modernization

Unfortunately, much like the history of military wargaming goes back centuries, so do most of the tools and techniques still used today to conduct modern wargaming. While there is still a role for traditional analog wargaming tools—such as physical game boards, cards, and dice—there is increasing pressure to finally bring wargaming into the 21$^{st}$ Century. As the battlespace becomes ever-increasingly complex and information-saturated, it is vital that we begin to leverage technologies, such as modeling and simulation (M&S), as well as modern advances, such as artificial intelligence (AI), "to evolve the current paradigm of wargaming—both in terms of technology and methodology" [3]. Today's emerging technologies such as distributed cloud computing, quantum computing, and machine learning could allow for the breakthroughs necessary to not only leverage machines for strategic, operational, and tactical decision aids, but also begin automating even the most complex tasks and decision-making that could only previously be done by humans—*not to replace humans, but to augment, inform, and ultimately improve human decision-making*.

Over the past decade, the United States (U.S.) Department of Defense (DOD) has seen a growing trend its polices, guidance, and reports which have highlighted the importance of sustaining, evolving, and even transforming our use of wargames and wargaming in support of force design, Professional Military Education (PME), concept development, and capability analysis [11]–[18]. Despite the former Deputy Secretary of Defense, Mr. Robert Work [17], calling for a reinvigoration of wargaming across the DOD back in 2015, little progress has been made in advancing, evolving, or transforming how the different Services have conducted wargames. In the 38$^{th}$ Commandant's Planning Guidance [12], General David Berger reiterated this need for an invigorated approach to wargaming and emphasized the need to invest more robustly in wargaming, experimentation, and M&S to allow us to better think, innovate, and change. Furthermore, General Berger recognized that although we have invested substantial energies into developing new concepts over the past two decades, we have done little to examine these concepts through rigorous wargaming, experimentation, and analysis.

In 2022, the Office of the Undersecretary of Defense for Research and Engineering (OUSD(R&E)) issued a report, *Technology Vision for an Era of Competition* [19], that called for the DOD to nurture early research and discover new scientific breakthroughs to prevent technological surprise. Furthermore, this report emphasized that our current era of strategic competition demands collective cooperation between the Government, the Defense Industrial Base (DIB), academia, Federally Funded Research and Development Centers (FFRDCs), University Affiliated Research Centers (UARCs), small businesses, and international partners. It emphasized that effective competition requires agility in initiating new technology development, rapid experimentation in relevant mission environments, and rapid transition to the users. Lastly, the report highlighted the need to adopt technologies already existent in the commercial sector in areas such as trusted

AI, autonomy, integrated network systems-of-systems, space technology, advanced computing and software, and human-machine interfaces.

While the need to modernize wargaming has been evident for some time, leveraging and adopting technologies from the commercial sector is a pivotal aspect of this transformation. The Final Report of the Defense Science Board (DSB) Task Force on Gaming, Exercising, Modeling, and Simulation (GEMS) [20] highlighted how how advancements in computing power, graphics, and other technologies have significantly enhanced GEMS capabilities—providing the DOD with cost-effective and innovative ways to test new ideas and concepts, design and prototype new systems, model military campaigns, conduct geopolitical analysis, and provide training to improve warfighter readiness and performance. The DSB further reiterated the importance of leveraging GEMS technologies if the DOD is to truly address the future challenges in training, systems development, acquisition, deterrence, and warfighting [20].

## 1.2 The People's Republic of China Wargaming Modernization

While the DOD has been relatively slow in making substantial progress in applying emerging technologies to advancing how we plan, execute, and analyze wargames, our adversaries, on the other hand, have swiftly grasped onto this opportunity to leverage these advancements and transform how they conduct and leverage wargaming for everything from improving military planning, to enhancing PME, to refining command and control. Even more concerning is that the People's Liberation Army's (PLA's) investment in AI is now on par with the U.S. DOD's AI spending [21].

Within the People's Republic of China (PRC), the history of wargaming as part of military planning is a well-established tradition [22] and has a long history that dates back to Sun Tzu [23]. Furthermore, given PLA's self-acknowledged relative lack of recent real-world combat experience, they have increasingly begun turning to wargaming to help bridge this experience gap between itself and its global competitors [22]. As the PLA began to train and educate their officers to become more familiar with modern technology and the increasing complexity of modern warfare, they have realized that the traditional classroom discussions are no longer sufficient [23]. This resulted in the creation of computerized wargaming and other M&S tools that started as far back as the 1990s [23].

Initially, these tools were distrusted by leadership and seen as clunky and uncoordinated. However, in 2007, the PLA renewed their efforts in developing better computerized wargaming systems to familiarize their commanders with decision-making under dynamic and constantly changing conditions [23]. In this modern era, the PLA saw the traditional emphasis on process and procedures quickly becoming outdated and realized the importance of accelerating the decision-making cycle to keep up with the "informationized" battlefield that no longer allows for delays and often requires immediate responses to developing circumstances [23]. Ultimately, the PLA sees computerized wargames as promoting more flexible tactics and closer situational analysis, while also training commanders and staffs to "think more completely, more precisely, more deeply, which will produce more effective levels of command stratagem" [23]. Thus, the PLA is now seeking to incorporate more realism in wargames to help emulate both the uncertainty and the time urgency present in modern warfare.

In fact, the PLA have prioritized computerized approaches over more traditional forms [7]. And as called out in the DSB's report [24] regarding our own need to leverage cost-effective techniques for training, the PLA has already begun leveraging these cost-effective techniques in their own training in an effort to address some of their long-standing weaknesses—such as command decision-making—and have also scaled up their wargaming efforts at their own PME institutions [7], [25].

Thus, wargaming has become more popular and prominent across the PRC as a whole, to the point where the PLA has leveraged the commercialization of wargaming as well as advances and innovations from the video game industry to provide its military forces with high-end, realistic, and engaging wargames [7]. In fact, the PRC has relied on its national strategy for military-civil fusion to partner their PLA with technology companies to advance wargaming and military simulations [7]. What is more, wargaming has now gone beyond the PLA's PME system and has been further extended to become a critical element in their national defense education where thousands of both military and civilian students at universities nationwide participate in annual wargaming competitions [7], [25].

It should be of no surprise then that the PLA is aggressively pursuing innovation in both the platforms and techniques used in wargaming [7]. One such innovations is the introduction of AI for wargaming [7], [22], [26] which falls under the PLA's "intelligentization" priority. In fact, the PLA saw the success of Google's AlphaGo's victory over the world-champion Go Master Lee Sedol, and AI's success in Texas Hold 'em poker as proof that AI could be applied to wargaming [7]. One specific example of their emerging capabilities is an AI-based wargaming simulation called *AlphaWar* which was inspired by DeepMind's AlphaStar AI system that resulted in a superhuman-AI player for StarCraft II [22].

If this were not enough, the PLA is also leveraging wargaming platforms to more quickly advance technological experimentation [7]. Recent progress includes PLA contests and competitions that have concentrated on developing AI systems for wargaming in complex scenarios [7]. The PLA is leveraging this human-machine confrontation to find ways to improve planning and decision support tools for future joint operations while also improving adversary behaviors in simulation [7], [22], [25]. One such example of an AI-assisted system developed by the PRC is the *Mozi Joint Operations Deduction System* that they have begun using in training and education [7]. *Mozi* is a human-in-the-loop deduction system for multi-domain joint operations capable of supporting scenarios from the campaign levels down to the tactical levels on the entire process of combat organization, planning, and command and control [7]. In fact, this system was inspired by and meant to imitate *Command: Modern Air Naval Operations* (CMANO), a commercial-off-the-shelf wargaming platform used by the U.S. and many NATO partners [7]. Relative to previous PLA wargames, *Mozi* allows for significant improvements while also serving as a platform designed to create and train AI agents [7].

## 1.3 Leveraging Artificial Intelligence for Wargaming

Given these concerning activities and the recent breakthroughs in AI's transformative capabilities, it becomes clear that we must begin investing more seriously in the development of AI for wargaming specifically. The National Security Commission on Artificial Intelligence (NSCAI) [27] details two convictions: (1) "the rapidly improving ability of computer systems to solve problems and to perform tasks that would otherwise require human intelligence—and in some instances exceed human performance—is world altering;" and (2) "AI is expanding the window of vulnerability the United States has already entered." Thus, given these, the NSCAI concludes that "the United States must act now to field AI systems and invest substantially more resources in AI innovation to protect its security, promote prosperity, and safeguard the future of democracy" [27]. By advancing AI-enabled systems and pairing these with human judgment, the NSCAI [27] contends that we will be able to enhance all-domain awareness, improve the speed and quality of our decision cycles, offer recommendations for different COAs, and more rapidly counter our adversary's actions.

Although the U.S. has enjoyed military superiority in most domains, the democratization of machine learning (ML) has begun to provide our competitors and other state actors with innumerable opportunities for disruption [28]. Thus, more than ever before, it becomes imperative that we aggressively research and experiment to build a solid ground-level understanding of the strengths and weaknesses of AI and how it can be used for planning and wargaming—only then can the DOD be better prepared and postured to deal with strategic surprise and disruption [28]. For example, COA analysis today largely focuses on assessing friendly plans with very little emphasis on how an adversary might react based on their own objectives and capabilities [26]. Although we go to great lengths at trying to understand our adversary's thinking and how they would act during a conflict, we will always be limited by our own imagination. Thomas Schelling puts it best in his Impossibility Theorem: "One thing a person cannot do, no matter how rigorous his analysis or heroic imagination, is to draw up a list of things that would not occur to him" [29]. AI-enabled wargaming could potentially allow us to overcome even this limitation by creating intelligent agents with their own goals who may not necessarily be constrained to our way of thinking and planning which is typically ingrained in us through sometimes decades of experience. Furthermore, learning new behaviors from data alone, AI provides us with the ability to automatically perform tasks that would otherwise require human intelligence [30].

In a report funded by the United Kingdom (UK) Defence Science and Technology Laboratory (Dstl), the Centre for Emerging Technology and Security (CETaS) conducted a study, *Artificial Intelligence in Wargaming: An evidence-based assessment of AI applications* [31], which involved a literature review, expert interviews, case study analyses, and a workshop convening experts across the defense and game AI communities. In their study, Knack and Rosamund point out that better understanding and appropriately employing AI could result in a "potential revolutionary change in wargaming" and—by investing in AI-enabling technologies—we may be able to achieve a decision advantage over our adversaries by introducing novel data analytics techniques for decision-makers while also facilitating the analysis of these same decisions.

In this report, Knack and Rosamund [31] identified a list of use cases consolidated across a group of professional wargame designers, M&S, and non-defense AI experts. These uses cases spanned different elements of wargame design, wargame execution, wargame analysis, and wargame logistics. Areas identified included those which could take advantage of AI solutions in existence today currently being used for other applications—such as automatic speech transcription—to more revolutionary high-risk, high-reward AI solutions to support COA generation and adjudication [31].

While substantial research has gone into advancing the field of ML, wargaming and military planning are significantly different than the traditional problems we have been using AI to solve thus far—such as image classification and natural language processing. Mission analysis and planning generally require human intuition and heuristics to be applied early-on to limit the size of the search problem [28]. While heuristics do allow us to find acceptable solutions more easily, these solutions are typically not scalable or reliable enough to evaluate the vast number of contingencies that might arise [28]. Furthermore, intuition can also fail in very complex problems, such as those involving high-dimensional spaces with many different players and complex weapons and sensor interactions [28]. Unfortunately, these complexities are the very characteristics that may define the future of warfare [26], [28].

To date, fortunately, competitive games have served as a good test-bed for learning how to implement AI in support of wargaming. Early successes have included mastering checkers [32], backgammon [33], chess [34], and Go [35]. AI methods have also achieved success in videos games such as Atari games [36], Super Mario Bros [37], Quake III [38], Dota 2 [39], StarCraft II [40], and No-Limit Texas Hold 'Em Poker [41]. However, competitive games are usually characterized by a set of fixed rules, defined parameters, and predictable outcomes based on known variables. While these games offer valuable insights into strategy, decision-making, and risk assessment, real-world wargaming scenarios are often more complex—with more possible initial game states and a larger branching factor—resulting in more unpredictable outcomes. This makes the translation of AI success from these games to genuine military operations a challenge. Nevertheless, the advancements in AI learning and adaptability garnered from these games provide a strong foundation for more nuanced applications of AI in combat simulations.

## 1.4    Leveraging the "Centaur" Concept for Wargaming

As the CeTAS report [31] details, there exist a plethora of different ways we can employ AI in support of wargaming; however, to scope the remainder of this paper, we will discuss AI as it relates to wargaming within the context of creating intelligent agents capable of making rational decisions despite the large and complex state spaces characteristic of combat modeling and simulation.

Showing that an AI can win games or achieve superhuman performance, however, is just the first step to showing that AI can actually provide useful insight to wargamers, operational planners, and battlefield commanders [42]. Nevertheless, we envision these intelligent agents serving as the foundation for creating modern decision-aid tools that can provide decision-makers more accuracy, speed, and agility over the more traditional tools [28]—potentially speeding up our decision-making process and offering pivotal insights. Neglecting this step, we believe, poses significant risks as we delve further into multi-domain operations [26] against an AI-enabled adversary.

While the idea of human-computer collaboration was originally envisioned by Licklider in 1960 [43], former world champion chess player Gary Kasparov first introduced the idea of "Centaur Chess"—where a human collaborates with a computer during play—years after his loss to IBM's Deep Blue in 1997 [44]. Despite having been defeated by an AI, Kasparov, instead of seeing AI as a threat, encouraged viewing AI as a tool that—when combined with human capabilities—can lead to unprecedented achievements [44]. In his book *Deep Thinking: Where Machine Intelligence Ends and Human Creativity Begins* [44], Kasparov highlighted the need to leverage the complementary strengths humans and machines have. Computers excel in brute-force calculations, able to analyze millions of positions per second while easily calculating the best near-term tactical move. Humans, on the other hand, have a much deeper understanding of strategy, creativity, and the ability to consider the long-term implications of a particular move—all while using mostly intuition [44]. Kasparov argued that the combined strength of this human intuition and the machine's calculations often led to stronger play than either top grandmasters or computers alone. In many instances, Kasparov noted, even relatively lower-ranked players paired with computers could outperform top-tier grandmasters.

Interestingly, Kasparov also pointed out that as computer chess programs became stronger, the role of the human player in this centaur partnership also evolved. While initially the humans would focus on strategy while the computers focused on tactics, as chess AIs improved, the humans increasingly began playing the role of "quality control," ensuring that the computer's recommended moves aligned with the human's broader strategic goals [44]. In fact, Kasparov often remarked that the future of chess might not be about humans versus machines, but about which humans, paired with which machines, using what interfaces, can play the best. This collaboration becomes a blend of the machine's computational strength and the human's ability to provide context, understanding, and intuition—and this synergy has produced levels of play greater than either could achieve on their own.

## 2.0 DEVELOPING ARTIFICIAL INTELLIGENCE FOR WARGAMING

Although there exist many different AI techniques and approaches that could be applied to wargaming such as supervised learning, unsupervised learning, genetic algorithms, natural language processing, decision trees, expert systems, game theory, adversarial networks, and many others, this paper focuses mainly on the need to advance the field of reinforcement learning (RL) in support of developing intelligent agent behaviors for wargaming.

When it comes to machine learning, there mainly exist three main types: supervised, unsupervised, and reinforcement learning. Supervised learning relies on labeled data where each set of inputs has a corresponding desired output. It is akin to learning by example and is best suited for tasks like image classification, regression, and speech recognition. Unsupervised learning, conversely, does not rely on labeled data. Instead, it finds patterns or structures within the data—such as grouping or clustering data points—and is best suited for anomaly detection, dimensionality reduction, and data segmentation. Of note, there also exists other types of machine learning such as transfer learning, active learning, self-supervised, learning, etc.; however, these are typically extensions or combinations of the two categories listed above.

## 2.1 Reinforcement Learning

The third main category, reinforcement learning (RL), is a type of machine learning where an agent learns to make decisions by performing certain actions in an environment to maximize some notion of cumulative reward. The agent learns from trial and error, receiving rewards or penalties based on the actions it takes, with the goal of finding the best strategy, or policy, to achieve the highest reward over time. Moreover, actions can not only affect the immediate rewards, but also follow-on rewards. In other words, RL is able to handle sparse and delayed rewards. These two elements together—trial-and-error search and delayed rewards—are what distinguishes RL from other types of machine learning.

More formally put, a reinforcement learning problem, shown in Figure 1, typically consists of a decision-maker, referred to as an *agent*, and an *environment*, represented by *states* $s \in S$. The *agent* can take *actions* $a_t$ as a function of the current state $s_t$ such that $a_t \in A(s_t)$. After choosing an *action* at time $t(a_t)$, the agent first receives a *reward* $r_{t+1}$ and finds itself in a new *state* $s_{t+1}$. The *action* $a_t$ comes from a strategy called a *policy* $\pi$. This *policy* $\pi$ is a mapping from *states* $s \in S$ to a probability of selecting each possible action $\pi(s,a)$. As the agent interacts with the environment, it will learn the optimal policy that maximizes its reward in the long run.

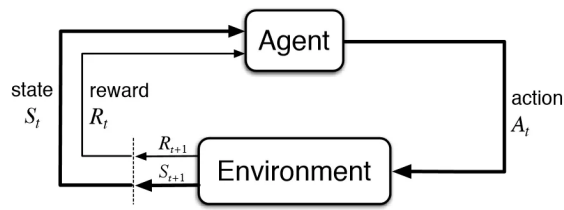

**Figure 1-1: Reinforcement Learning**

Early applications of RL involved table-based methods, such as Q-learning, to address problems with limited state and action spaces [45]. Traditional Q-learning maintains a table where each cell symbolizes the value of a state-action pair, denoted as $Q(s, a)$. While straightforward and capable of finding an optimal policy for small state spaces with consistent learning and exploration, the sheer growth of state or action spaces can quickly make for an intractable problem to search through exhaustively. The game of chess, for example, has a vast number of potential state-action combinations, with a game-tree complexity of $10^{120}$ [46]. For comparison, the number of atoms in the universe is broadly estimated to be $10^{80}$ [47]—an exponentially smaller number than that of the complexity of chess. Needless to say, the complexity of realistic combat simulation scenarios is orders of magnitude greater than that of the rule-bound game of chess.

Fortunately, recent advances in neural networks have given rise to deep reinforcement learning (DRL) [36]. DRL integrates deep learning techniques with RL, where deep neural networks are used to approximate value functions or policies, enabling the handling of large-scale and high-dimensional environments for which table-based methods are impractical. Instead of explicit storage of each state-action value, deep neural networks can predict these Q-values with just the input state. This allows the learner to better generalize from familiar or known states to novel states—resulting in better predictions even in less-explored areas of the state space. Additionally, deep neural networks can provide for a more memory-efficient, compact representation for vast state spaces and can handle even more complex tasks due to their capacity for representing complex non-linear functions. Thus, algorithms such as Deep Q-Networks (DQN) [36], which leverages deep neural networks for Q-value estimation, have inspired other DRL algorithms such as policy gradient methods [48], Actor-Critic [49], and Asynchronous Advantage Actor-Critic (A3C) [50].

However, despite the integration of deep neural networks with RL having solved some problems, it has also introduced its own set of challenges. Due to the non-stationary nature of data, RL can sometime result in unstable learning. Furthermore, as with any deep learning approach, there is always the potential risk of models overfitting to specific states or trajectories which could lead them to underperforming or acting irrationally in more novel states. Additionally, DRL is known to be sample inefficient, requiring vast amounts of data to learn effectively [51]. Nevertheless, to address these challenges, there have been strategies developed within the last decade such as Experience Replay [52] and Target Networks [53] to stabilize learning, and diverse exploration strategies such as entropy-regularization to reduce overfitting [54]. This continued evolution of DRL has allowed for more broad applicability for RL to address more complex, high-dimensional tasks than has previously been possible.

### 2.1.1 Markov Decision Process

Within RL, the concept of a Markov Decision Process (MDP) is the foundational mechanism for sequential decision making [55]. The *Markov* property itself simply states that the future only depends on the present, and not the past. MDPs allow for a mathematical representation of the RL problem [45], as well as provide an underlying theory in how to maximize expected rewards over time. This model consists of choosing an action in a state which then generates a reward and determines the next state through a transition probability function. Integral to MDP's mathematical representation is the *Discount Factor* $\gamma$ which determines the present value of future rewards. For wargaming, this could represent strategic foresight and the delicate balance commanders and decision-makers must consider between sacrificing immediate gains for more significant future benefits, or vice versa.

Additionally, the recursive nature of decision-making in RL, mathematically captured by the Bellman Equation [56], is similar to the iterative nature of decision-making in wargaming where each tactical choice, influenced by prior decisions, impacts the subsequent state of play and actions. Furthermore, the real-world challenges of wargaming can be mirrored in the Partially Observable MDPs (POMDPs) [57]. Needless to say, all actions in war involve some level of uncertainty, or "fog of war" [58], where critical information about the enemy, environment, or friendly forces is uncertain or unknown. POMDPs allow for representing even this level of complexity—providing for a way to simulate the uncertainties planners and decision-makers must factor into their own plans. Moreover, transition dynamics within MDPs capture the inherent unpredictability of warfare. Thus, as players refine their strategies, the underlying MDP can adapt in real-time, ensuring a continually evolving challenge.

Ultimately, the MDP framework is what enables adaptive opponents and offers scalability through approximation methods of the value or policy functions, ensuring that DRL remains relevant even in the most intricate wargaming setups. In essence, MDPs go beyond simply providing a mathematical foundation for RL—it offers a way to model the complexities, challenges, and dynamics of warfare.

### 2.1.2 Applying RL to Wargaming

It should become clear then how RL provides for one way to apply AI to wargaming that has the potential for a multitude of transformative benefits. Unlike traditional AI approaches that have been applied to combat M&S thus far—which are often limited to fixed algorithms or scripted behaviors—RL opens the door to creating agents that can adaptively learn from their environments. This allows an RL-based AI to continually refine its tactics and strategies based on evolving wargaming scenarios and human (or AI-based) opponents. Furthermore, RL may be more effective at addressing the unpredictable nature of warfare—where decisions may have long-term consequences—better than other AI approaches.

This offers wargamers, planners, and decision-makers not just a more dynamic and challenging adversary, but also aids in better decision-making and operational planning. Through RL's adaptive reward systems, agents can be fine-tuned to achieve particular strategic or tactical objectives. For instance, if a wargaming scenario calls for preserving forces, it can be adapted to prioritize this objective. Additionally, RL-trained adversary agents may be more capable of pushing participants to their maximum capabilities, exposing potential weaknesses and promoting constant skill development. This ability of RL agents to continually learn and improve also allows them to adapt to emerging warfighting concepts and discover potentially novel ways of exploiting doctrinal or newly developed strategies and tactics. Through RL, we can either replicate realistic adversary strategies or generate new optimized strategies and tactics entirely—introducing players to novel strategic challenges they might not encounter when playing against human counterparts.

One of RL's most understated strengths is its ability to handle the inherent unpredictability of war and complement human intuition. The tactics an RL agent may develop in simulation could be an untapped source for truly "outside-the-box" strategic insights. Often, an RL-driven perspective can unveil innovative approaches or tactics that may be overlooked by experienced human planners and strategists. Kasparov noted a similar effect where human players learned new tactics from the chess AIs that a millennia of human chess playing had not yet discovered [44]. Shin et al. [59] also noted this phenomena of improved decision quality when human Go players had access to the reasoning process of the AlphaGo AI as compared to those who did not have access to AI.

It must be acknowledged then, that human judgment, while absolutely invaluable, can sometimes be clouded by biases or narrow-sightedness based on our individual and collective experiences, training, and education. Most, if not all, senior military planners, decision-makers, and commanders have undergone the same military training, PME, and career experiences, resulting in a high likelihood of having developed homogenous ways of thinking and approaches to problem-solving. An RL agent, on the other hand—free from such experiential biases—can offer fresh, "outside-the-box" solutions and viewpoints that may complement our expert, human-derived strategies and tactics.

This idea of leveraging RL for gameplaying for wargaming is not new, however. Therefore, as we continue to advance the study and application of AI in this domain, it is vital we build upon relevant past and ongoing research. Key RL research that can be leveraged for applying AI to wargaming include intelligent game-playing agents created for Go [35], StarCraft II [40], and Dota 2 [39]. Research that has explored using RL to develop intelligent agents for wargaming include research by Alt [60], Kurniawan et al. [61], Toghiani-Rizi et al. [62], Zhang et al. [63], Boron and Darken [64], Yu et al. [65], Cannon and Goericke [66], Coble [67], Finley [68], and Allen [69].

## 2.2 Hierarchical Reinforcement Learning

Despite RL having demonstrated many successful use-cases to date in achieving human- or superhuman-level performances in smaller and more constrained scenarios or games, RL agents still need to be able to appropriately scale to meet the requirements of large wargames involving hundreds or even thousands of entities and large state spaces. As the different types and number of entities increase in a wargame, however, so does the amount of information (i.e., state and action spaces)—quickly becoming an intractable problem. Unfortunately, research to date leveraging RL to model intelligent agent behaviors have thus far struggled to generalize to larger and more complex scenarios [8], [66].

To address these more complex combat simulation scenarios, we turn to hierarchical reinforcement learning (HRL)—an extension to RL which adds a hierarchical dimension in an attempt to simplify the problem. Hierarchical decomposition is a natural way humans break down complex tasks and re-use old or related skills [70]. RL, on the other hand, must master hundreds or even thousands of small tasks independently and from scratch, which—though possible for relatively simple games—is computationally very expensive and often intractable for more complex games.

HRL shows promise to help address this problem, also known as Bellman's "curse of dimensionality," in that it "decomposes a reinforcement learning problem into a hierarchy of subproblems or subtasks such that higher-level parent-tasks invoke lower-level child tasks as if they were primitive actions" [71]. This decomposition may itself have multiple levels of hierarchy, of which some or all could be RL problems themselves. By decomposing the problem using hierarchical models, we may be able to solve more complex problems by reducing the computational complexity given that the overall problem can be appropriately represented more compactly, and their subtasks be reused or learned independently. Of note, however, given the constraints of the hierarchy, although the solution to an HRL problem may be optimal, there is no guarantee that the decomposed solution is an optimal solution to the original problem [71].

Nevertheless, using a hierarchical decomposition might facilitate the generation of complex behaviors by attempting to solve the problem at multiple levels of abstraction [72]. In fact, HRL has already shown to solve very large-scale problems involving state spaces of $10^{100}$ with branching factors of $10^{30}$ [72]. This decomposition may improve the time and space complexity in both the learning and execution aspects of RL [71].

In their chapter *Behavioral Hierarchy: Exploration and Representation*, Barto et al. [73] discuss how behavioral modules can provide reusable building blocks that can be composed hierarchically to generate an extensive range of behaviors. Of interest, they list four primary benefits of employing behavioral hierarchies. The first, and most beneficial, aspect of leveraging behavioral hierarchies is that higher-level actions can make use of lower-level actions without concern for execution specifics. This enables learning complex skills and planning at different levels of abstraction [73]. Agents can now create hierarchies of behavior modules (i.e., skills), effectively decreasing the search space since selecting between different higher-level skills allows the agent to take larger and more meaningful steps through the search space—as opposed to having to search each individual possible primitive action. The second benefit for behavior hierarchies is that it naturally enables transfer learning. The behavior modularity, or "portable skills" enables the transfer of results of learning from one task to another related task, rather than continually trying to acquire new skills from scratch. A third benefit is the added exploratory abilities an agent can gain from a hierarchical approach [73]. New skills learned by an agent become atomic behavioral modules that can be used when computing higher-level behavior strategies and learning more complex skills. This growing set of skills allows an agent to reach farther into the state space than may have previously been inaccessible, which in turn enables new skill discovery and allows for discovering more complex environmental dynamics. The fourth and last benefit of behavioral hierarchies is its ability to reduce the representational challenges of learning (or planning) in large, complex domains. Hierarchies allow each module to only contain its module-specific representation that only includes what is needed for its specific operation while excluding irrelevant data [73]. This can ultimately make learning feasible despite very large state and action spaces.

Key to HRL and the mechanism to apply the behavioral hierarchies discussed above is the concept of *options* as introduced by Sutton et al. [74]. *Options* are a generalization of *actions*, which Sutton et al. formally used only for primitive choices. Previous terms have included *macro-actions*, *behaviors*, *abstract actions*, and *subcontrollers* [74]. While a *primitive action* typically affects the environment only for a single time step, an *option* can persist over multiple time steps. This concept allows for the decomposition of tasks into temporally extended courses of action as proposed by Barto et al. [73].

Thus, *options* provide a framework to abstract and encapsulate frequently used sequences of actions into higher-level behaviors, or what we term *decisions*. This has the potential to reduce the complexity of learning in that, instead of learning every step for a complex task, the agent can learn to use a sequence of more abstract *options*, each of which represents a simpler sequence of subtasks (or sequence of *primitive actions*). Additionally, the *options* or HRL framework may allow for improved exploration of the state space at different levels of abstractions. Rather than having to explore only one step at a time, *options* potentially allows an agent to explore multiple timesteps away. Lastly, this framework can facilitate transfer learning in that, once an *option* is learned, it can be used across multiple tasks. For example, if an agent learns the *option* of executing a flanking attack, it can potentially be recalled across different types of scenarios; whereas only relying on traditional RL would likely require training an agent across very similar scenarios for which it would have to independently learn the flanking attack.

The idea of using HRL to simplify the RL problem is also not new. Key RL research that has advanced the science of applying HRL to domains that wargaming can benefit from include research from Dayan and Hinton [75], Vezhnevets et al. [76], Levy et al. [77], Frans et al. [70], Florensa et al. [78], Pope et al. [79], Wang et al. [80], Zhang et al.[81], Li et al. [82], Wang et al. [83], and Rood [8].

## 3.0 SCALING ARTIFICIAL INTELLIGENCE FOR WARGAMING VIA HIERARCHICAL REINFORCEMENT LEARNING

To address the challenges inherent in scaling AI to deal with the complexities of warfare, we are leveraging recent successes in RL and the nascent research in HRL to develop unscripted, intelligent agents capable of operating in large, complex environments while limiting the number of training episodes to a reasonable amount (e.g., limiting the training to that which could be conducted on a DOD-accessible high-performance computer over the period of a few days). We recognize that this is only one use-case on how AI can be used in support of wargaming. Nevertheless, we believe that the integration of intelligent agents in combat simulations is a foundational element necessary for AI-based COA generation and analysis—as well as for the development of advanced human-machine teaming decision aids in the near future.

To accomplish this goal, we are addressing the current challenges in applying AI to wargaming by:

1. Developing an agent architecture consisting of a hierarchical decomposition of agents, decisions, and policies to manage the exponential growth in computations required by large state-action spaces.

2. Developing an HRL training framework that allows for training multiple hierarchical layers of agents, decisions, and policies using different levels of observation abstractions.

3. Developing independent, hierarchical, dimension-invariant observation abstractions for each level of the hierarchy that will further enable both scalability and re-use of RL models.

4. Developing a multi-model approach that can dynamically switch between different independently-trained behavior models (e.g., machine learning, expert systems, rule-based, optimization algorithms, game solving) allowing us to leverage very specialized models rather than attempting to train a single model capable of optimal performance in any situation.

5. Demonstrating the scalability of our framework and architecture by implementing in both a low-fidelity and a high-fidelity combat simulation.

Our agent architecture consists of developing an embedded hierarchy of agents, decisions, and policies. The agent hierarchy primarily decomposes the force into groups of 3 to 5 units, where each group is controlled by a *manager* agent. These *manager* agents are then grouped together and controlled by a *commander* agent. Simple scenarios with few units may require only two levels of agents, whereas complex scenarios involving dozens of units could benefit from three or more levels of hierarchy. Because our research intends to examine more complex scenarios, we anticipate needing at least three levels of hierarchy.

Within the hierarchy of agents is a hierarchy of decisions. Having multiple levels for decisions allows each level to be trained towards different goals and at different levels of abstraction—making scaling to very complex scenarios a more tractable problem. Additionally, this hierarchical approach implicitly trains for agent coordination and cooperation since the layer above controls the aggregate-level behaviors of the layers below. With the exception of the bottom-most level, each level of the decision hierarchy can be thought of as being abstract or cognitive (i.e., they are initially high-level decisions that will ultimately inform a primitive action). Only the agent at the bottom of the hierarchy is an actual entity on the gameboard who takes a discrete or primitive action that directly affects the environment.

To date, we have found that creating dimension-invariant observation abstraction have allowed us to train agents on larger gameboards—while still producing intelligent behavior—than has previously been possible given a reasonable training budget. We use different levels of abstractions based on the level of the hierarchy being computed. Commanders at the highest level of the hierarchy use a coarse abstraction of the gameboard, while the actual units used in the lowest-level of the hierarchy use a higher-fidelity, yet localized, abstraction of the state space. In this specific unit-level abstraction, rather than taking the full state of the game, we create a localized abstraction centered on the agent on-move (i.e., the unit on the gameboard whose turn it is to take an action). We then use a piecewise linear spatial decay function to calculate weights based on how far away the item of interest is from the unit on-move. These weights are then multiplied by the values within each respective cell of each channel, summed up by radial, and then inserted into the outermost cell layer of the respective channels in the agents localized observation space. This provides the agent with an orthogonal representation of information up to a finite number of hexagons away (e.g., 3 hexagons), but still have decayed information represented of items that might be farther away and thus currently less relevant to the agent.

We have also found success in a multi-model framework we developed where, at each action-selection step, the multi-model takes in an observation as input, and passes it to each of its score prediction models. Each score prediction model then infers a predicted game score, which is fed into an evaluation function. A specific behavior model is then selected based on this evaluation function. Finally, the original observation is passed to the selected behavior model, which then produces an action. To supply the evaluation function that selects the appropriate behavior model, we train a separate score prediction model for each individual behavior model in the repository. This score prediction model is a convolutional neural network (CNN) which infers a game score based on the current game state. This predicted game score assumes that the blue faction continues playing the game according to their respective behavior model, and the red faction continues playing according to a specific adversary behavior model. Given that the combat simulation used for our research, Atlatl [84], is a turn-based game and not a time-step simulation, we refer to each instance where an entity on the gameboard is prompted to take an action as an action-selection step. Although to date we have trained our score prediction models using supervised learning with game data, we have recently developed and are currently testing a separate version of this score prediction model that leverages DRL instead of supervised learning—potentially enabling easier scalability of this approach.

Our next step is to develop the second level, the manager level, of our HRL framework that will supply higher-level goals to our individual units on the board. Following this, we will develop the third level, or commander level, of our HRL framework. Integral to this portion of the research will be to determine the specific inputs and outputs of the higher-level agents. We intend to continue experimenting and determine which levels of abstraction are appropriate based on the decisions to be made at each level of the hierarchy—from the operational level down to the tactical level. Additionally, we will continue experimenting with goal and subgoal inputs and outputs as well as the appropriate rewards engineering necessary to produce intelligent, rational decisions by the higher-level agents. Along the way, we will also continue to refine each component of the HRL framework while refining and evolving our multi-model approach.

## 4.0  CONCLUSION

Military wargaming, a centuries-old tradition, still largely relies on age-old methods despite modern advances in technologies, such as M&S and AI. While the U.S. DOD acknowledges the need to modernize wargaming, progress remains slow, though various organizations are beginning to acknowledge the promise of emerging technologies. As highlighted by various reports and parallels from games that have benefited from AI thus far, such as chess and Go, AI will soon become critical to support and enhance decision-making for our planners and commanders across the force. To truly benefit from these advances, however, collaborations across sectors and embracing rapid technological transitions are vital. Moreover, adversaries like the PRC appear to have integrated modern technologies into their training, experimentation, and analysis, with the PLA already leveraging AI and commercial gaming advancements for enhanced wargaming.

More importantly, this fusion of machine learning with wargaming has the potential to go beyond just being a simple, incremental upgrade to our decision-making capabilities. It represents a potential paradigm shift, resulting in not just enhanced adversaries, but a holistic move towards achieving more adaptive and insightful decision aids. In today's modern battlefield, where technology can quickly reshape the dynamics of engagements and redefine strategic paradigms, RL-based intelligent agents not only can replicate doctrinal tactics and strategies, but also provide the ability to explore uncharted territories—potentially leading to novel insights previously unattainable through traditional planning methods. Thus, investing and developing AI within our M&S tools may allow us to leverage human intuition while also combining cognitive and computational capabilities necessary to unearth patterns, insights, or solutions that might remain hidden to our human decision-makers alone. In essence, intelligent agents in simulations will not only prepare us to tackle known difficult challenges but also equip us to potentially discover and exploit previously unrecognized opportunities.

Overall, our research to date has shown promise in this approach to scale the application of AI to more complex domains such as that of combat M&S in support of wargaming. Our multi-model framework has thus far allowed us to drastically improve the performance of our agents well beyond our current state-of-the-art scripted or RL-trained agents within our combat simulation. Our local observation abstraction of the state space shows potential in that, given a reasonable training budget, it now allows agents to learn on much bigger gameboards than have previously been possible within Atlatl specifically. We anticipate that incorporating these developments into our overall HRL framework will further improve the performance of our agents and potentially allow us to scale to virtually any size scenario.

## REFERENCES

[1] Joint Chiefs of Staff, "Joint Publication 5-0: Joint Planning." Dec. 01, 2020. [Online]. Available: https://irp.fas.org/doddir/dod/jp5_0.pdf

[2] F. J. McHugh, *U.S. Navy fundamentals of war gaming*. New York: Skyhorse Publishing, Inc., 2013.

[3] Y. Wong, S. Bae, E. Bartels, and B. Smith, *Next-Generation Wargaming for the U.S. Marine Corps: Recommended Courses of Action*. RAND Corporation, 2019. doi: 10.7249/RR2227.

[4] M. B. Caffrey, *On wargaming: how wargames have shaped history and how they may shape the future*. in Naval War College Newport papers, no. 43. Newport, Rhode Island: Naval War College Press, 2019.

[5] P. P. Perla, *The art of wargaming: a guide for professionals and hobbyists*. Annapolis, Md: Naval Institute Press, 1990.

[6] P. Perla, "Wargaming and The Cycle of Research and Learning," *Scand. J. Mil. Stud.*, vol. 5, no. 1, pp. 197–208, Sep. 2022, doi: 10.31374/sjms.124.

[7] E. B. Kania and I. B. McCaslin, "Learning Warfare from the Laboratory--China's Progression in Wargaming and Opposing Force Training," Institute for the Study of War, 2021. [Online]. Available: https://www.understandingwar.org/report/learning-warfare-laboratory-china's-progression-wargaming-and-opposing-force-training

[8] P. R. Rood, "Scaling Reinforcement Learning Through Feudal Muti-Agent Hierarchy," Naval Postgraduate School, Monterey, CA, 2022.

[9] P. A. G. Sabin, *Simulating war: studying conflict through simulation games*. London: Bloomsbury Academic, 2014.

[10] P. P. Perla and E. McGrady, "Why Wargaming Works," *Nav. War Coll. Rev.*, vol. 64, no. Number 3 Summer, p. 21, 2011, Accessed: Oct. 18, 2022. [Online]. Available: https://digital-

commons.usnwc.edu/cgi/viewcontent.cgi?article=1578&context=nwc-review

[11] Chief of Naval Operations, "Chief of Naval Operations Navigation Plan 2021." Jan. 2021. [Online]. Available: https://media.defense.gov/2021/Jan/11/2002562551/-1/-1/1/CNO%20NAVPLAN%202021%20-%20FINAL.PDF

[12] D. H. Berger, "38th Commandant of the Marine Corps Commandant's Planning Guidance." U.S. Marine Corps, 2020.

[13] D. H. Berger, "Force Design 2030." U.S. Marine Corps, May 09, 2022. [Online]. Available: https://www.hqmc.marines.mil/Portals/142/Docs/CMC38%20Force%20Design%202030%20Report%20Phase%20I%20and%20II.pdf?ver=2020-03-26-121328-460

[14] Department of the Navy, "Education for Seapower." Department of the Navy, Dec. 2018.

[15] Department of the Navy, "Education for Seapower Strategy." Department of the Navy, Feb. 28, 2020. [Online]. Available: https://media.defense.gov/2020/May/18/2002302033/-1/-1/1/NAVAL_EDUCATION_STRATEGY.PDF

[16] United States Government Accountability Office, "Additional Actions Could Enhance DOD's Wargaming Efforts," United States Government Accountability Office, Defense Analysis GAO-23-105351, Apr. 2023. [Online]. Available: https://www.gao.gov/assets/gao-23-105351.pdf

[17] Deputy Secretary of Defense, "Wargaming and Innovation." Feb. 09, 2015. [Online]. Available: https://news.usni.org/2015/03/18/document-memo-to-pentagon-leadership-on-wargaming

[18] D. H. Berger, "Training and Education 2030." U.S. Marine Corps, Jan. 2023. [Online]. Available: https://www.marines.mil/Portals/1/Docs/Training%20and%20Education%202030.pdf

[19] Office of the Under Secretary of Defense for Research and Engineering, "USD(R&E) Technology Vision for an Era of Competition." Feb. 01, 2022. [Online]. Available: https://www.cto.mil/wp-content/uploads/2022/02/usdre_strategic_vision_critical_tech_areas.pdf

[20] Defense Science Board, "Defense Science Board Gaming, Exercising, Modeling, and Simulation," Office of the Under Secretary of Defense for Research and Engineering, Jan. 2021.

[21] J. Harper, "China Matching Pentagon Spending on AI," *National Defense*, no. Robotics and Autonomous Systems, Jan. 06, 2022. Accessed: Nov. 15, 2022. [Online]. Available: https://www.nationaldefensemagazine.org/articles/2022/1/6/china-matching-pentagon-spending-on-ai

[22] R. Fedasiuk, J. Melot, and B. Murphy, "Harnessed Lightning: How the Chinese Military is Adopting Artificial Intelligence," Center for Security and Emerging Technology, Oct. 2021. doi: 10.51593/20200089.

[23] D. Cheng, "The People's Liberation Army on Wargaming," Feb. 17, 2015. https://warontherocks.com/2015/02/the-peoples-liberation-army-on-wargaming/ (accessed Oct. 18, 2022).

[24] "Defense Science Board GEMS Report," Office of the Under Secretary of Defense for Research and Engineering, Washington, D.C, Nov. 2020.

[25] E. B. Kania, "Artificial intelligence in China's revolution in military affairs," *J. Strateg. Stud.*, vol. 44, no. 4, pp. 515–542, Jun. 2021, doi: 10.1080/01402390.2021.1894136.

[26] P. Narayanan, M. Vindiola, S. Park, A. Logie, and N. Waytowich, "First-Year Report of ARL Director's Strategic Initiative (FY20–23): Artificial Intelligence (AI) for Command and Control (C2) of Multi-Domain Operations (MDO)," p. 32, May 2021.

[27] E. Schmidt *et al.*, "Final Report National Security Commission on Artificial Intelligence," National Security Commission on Artificial Intelligence, 2021. [Online]. Available: https://reports.nscai.gov/final-report/table-of-contents/

[28] L. A. Zhang *et al.*, *Air Dominance Through Machine Learning: A Preliminary Exploration of Artificial Intelligence–Assisted Mission Planning*. RAND Corporation, 2020. doi: 10.7249/RR4311.

[29] C. Turnitsa, C. Blais, and A. Tolk, Eds., *Simulation and Wargaming*. Hoboken, NJ: Wiley, 2021.

[30] C. Newton, J. Singleton, C. Copland, S. Kitchen, and J. Hudack, "Scalability in modeling and simulation systems for multi-agent, AI, and machine learning applications," in *Artificial Intelligence and Machine Learning for Multi-Domain Operations Applications III*, T. Pham, L. Solomon, and M. E. Hohil, Eds., Online Only, United States: SPIE, Apr. 2021, p. 70. doi: 10.1117/12.2585723.

[31] A. Knack and R. Powell, "Artificial Intelligence in Wargaming: An evidence-based assessment of AI

applications," *CETaS Res. Rep.*, Jun. 2023, [Online]. Available: https://cetas.turing.ac.uk/publications/artificial-intelligence-wargaming

[32] J. Schaeffer, J. Culberson, N. Treloar, B. Knight, P. Lu, and D. Szafron, "A world championship caliber checkers program," *Artif. Intell.*, vol. 53, no. 2–3, pp. 273–289, Feb. 1992, doi: 10.1016/0004-3702(92)90074-8.

[33] G. Tesauro, "TD-Gammon, a Self-Teaching Backgammon Program, Achieves Master-Level Play," *Neural Comput.*, vol. 6, no. 2, pp. 215–219, Mar. 1994, doi: 10.1162/neco.1994.6.2.215.

[34] M. Campbell, A. J. Hoane Jr., and F. Hsu, "Deep Blue," *Artif. Intell.*, vol. 134, no. 1–2, pp. 57–83, Jan. 2002, [Online]. Available: https://www.sciencedirect.com/journal/artificial-intelligence/vol/134/issue/1

[35] D. Silver et al., "Mastering Chess and Shogi by Self-Play with a General Reinforcement Learning Algorithm." arXiv, Dec. 05, 2017. Accessed: Sep. 07, 2022. [Online]. Available: http://arxiv.org/abs/1712.01815

[36] V. Mnih et al., "Human-Level Control Through Deep Reinforcement Learning," *Nature*, vol. 518, no. 7540, pp. 529–533, Feb. 2015, doi: 10.1038/nature14236.

[37] D. Pathak, P. Agrawal, A. A. Efros, and T. Darrell, "Curiosity-driven Exploration by Self-supervised Prediction." arXiv, May 15, 2017. Accessed: Sep. 07, 2022. [Online]. Available: http://arxiv.org/abs/1705.05363

[38] M. Jaderberg et al., "Human-level performance in 3D multiplayer games with population-based reinforcement learning," *Science*, vol. 364, no. 6443, pp. 859–865, May 2019, doi: 10.1126/science.aau6249.

[39] C. Berner et al., "Dota 2 with Large Scale Deep Reinforcement Learning," p. 66, 2019.

[40] O. Vinyals et al., "Grandmaster Level in StarCraft II Using Multi-Agent Reinforcement Learning," *Nature*, vol. 575, no. 7782, pp. 350–354, Nov. 2019, doi: 10.1038/s41586-019-1724-z.

[41] N. Brown and T. Sandholm, "Superhuman AI for heads-up no-limit poker: Libratus beats top professionals," *Science*, vol. 359, no. 6374, pp. 418–424, Jan. 2018, doi: 10.1126/science.aao1733.

[42] K. Atherton, "DARPA Wants Wargame AI To Never Fight Fair," *Breaknig Defense*, Aug. 18, 2020. [Online]. Available: https://breakingdefense.com/2020/08/darpa-wants-wargame-ai-to-never-fight-fair/

[43] J. C. R. Licklider, "Man-Computer Symbiosis," *IRE Trans. Hum. Factors Electron.*, vol. HFE-1, no. 1, pp. 4–11, Mar. 1960, doi: 10.1109/THFE2.1960.4503259.

[44] G. K. Kasparov and M. Greengard, *Deep thinking: where machine intelligence ends and human creativity begins*, First edition. New York: PublicAffairs, an imprint of Perseus Books, LLC, 2017.

[45] R. S. Sutton and A. G. Barto, *Reinforcement learning: an introduction*, Second edition. in Adaptive computation and machine learning series. Cambridge, Massachusetts: The MIT Press, 2018.

[46] C. E. Shannon, "Programming a computer for playing chess," *Lond. Edinb. Dublin Philos. Mag. J. Sci.*, vol. 41, no. 314, pp. 256–275, Mar. 1950, doi: 10.1080/14786445008521796.

[47] B. S. Ryden, *Introduction to cosmology*, Second edition. New York, NY: Cambridge University Press, 2017.

[48] J. Schulman, F. Wolski, P. Dhariwal, A. Radford, and O. Klimov, "Proximal Policy Optimization Algorithms." arXiv, Aug. 28, 2017. Accessed: Nov. 04, 2022. [Online]. Available: http://arxiv.org/abs/1707.06347

[49] T. Haarnoja et al., "Soft Actor-Critic Algorithms and Applications." arXiv, Jan. 29, 2019. Accessed: Nov. 27, 2022. [Online]. Available: http://arxiv.org/abs/1812.05905

[50] V. Mnih et al., "Asynchronous Methods for Deep Reinforcement Learning," p. 10, 2016.

[51] M. Botvinick, S. Ritter, J. X. Wang, Z. Kurth-Nelson, C. Blundell, and D. Hassabis, "Reinforcement Learning, Fast and Slow," *Trends Cogn. Sci.*, vol. 23, no. 5, pp. 408–422, May 2019, doi: 10.1016/j.tics.2019.02.006.

[52] V. Mnih et al., "Playing Atari with Deep Reinforcement Learning." arXiv, Dec. 19, 2013. Accessed: Sep. 10, 2023. [Online]. Available: http://arxiv.org/abs/1312.5602

[53] H. Van Hasselt, A. Guez, and D. Silver, "Deep Reinforcement Learning with Double Q-Learning," *Proc. AAAI Conf. Artif. Intell.*, vol. 30, no. 1, Mar. 2016, doi: 10.1609/aaai.v30i1.10295.

[54] T. Haarnoja, A. Zhou, P. Abbeel, and S. Levine, "Soft Actor-Critic: Off-Policy Maximum Entropy Deep Reinforcement Learning with a Stochastic Actor." arXiv, Aug. 08, 2018. Accessed: Sep. 07, 2022. [Online]. Available: http://arxiv.org/abs/1801.01290

[55] M. L. Puterman, *Markov Decision Processes: Discrete Stochastic Dynamic Programming*, 1st ed. in Wiley Series in Probability and Statistics. Wiley, 1994. doi: 10.1002/9780470316887.

[56] R. Bellman, "The Theory of Dynamic Programming," *Bull. Am. Math. Soc.*, vol. 60, pp. 503–515, Jul. 1954, [Online]. Available: https://apps.dtic.mil/sti/citations/AD0604386

[57] L. P. Kaelbling, M. L. Littman, and A. R. Cassandra, "Planning and acting in partially observable stochastic domains," *Artif. Intell.*, vol. 101, no. 1–2, pp. 99–134, May 1998, doi: 10.1016/S0004-3702(98)00023-X.

[58] *MCDP 1 Warfighting*. Washington, D.C: U.S. Marine Corps, 1989.

[59] M. Shin, J. Kim, and M. Kim, "Human Learning from Artificial Intelligence: Evidence from Human Go Players' Decisions after AlphaGo," in *Proceedings of the Annual Meeting of the Cognitive Science Society*, Vienna, Austria, Jul. 2021. doi: 10.5281/ZENODO.5214454.

[60] J. Alt, "Learning from Noisy and Delayed Rewards: The Value of Reinforcement Learning to Defense Modeling and Simulation," Naval Postgraduate School, Monterey, CA, 2012. [Online]. Available: https://apps.dtic.mil/sti/pdfs/ADA567384.pdf

[61] B. Kurniawan, P. Vamplew, M. Papasimeon, R. Dazeley, and C. Foale, "An Empirical Study of Reward Structures for Actor-Critic Reinforcement Learning in Air Combat Manoeuvring Simulation," in *AI 2019: Advances in Artificial Intelligence*, J. Liu and J. Bailey, Eds., in Lecture Notes in Computer Science, vol. 11919. Cham: Springer International Publishing, 2019, pp. 54–65. doi: 10.1007/978-3-030-35288-2_5.

[62] B. Toghiani-Rizi, F. Kamrani, L. J. Luotsinen, and L. Gisslen, "Evaluating deep reinforcement learning for computer generated forces in ground combat simulation," in *2017 IEEE International Conference on Systems, Man, and Cybernetics (SMC)*, Banff, AB: IEEE, Oct. 2017, pp. 3433–3438. doi: 10.1109/SMC.2017.8123161.

[63] G. Zhang, Y. Li, X. Xu, and H. Dai, "Efficient Training Techniques for Multi-Agent Reinforcement Learning in Combat Tasks," *IEEE Access*, vol. 7, pp. 109301–109310, 2019, doi: 10.1109/ACCESS.2019.2933454.

[64] J. Boron and C. Darken, "Developing Combat Behavior through Reinforcement Learning in Wargames and Simulations," in *2020 IEEE Conference on Games (CoG)*, Osaka, Japan: IEEE, Aug. 2020, pp. 728–731. doi: 10.1109/CoG47356.2020.9231609.

[65] S. Yu, W. Zhu, and Y. Wang, "Research on Wargame Decision-Making Method Based on Multi-Agent Deep Deterministic Policy Gradient," *Appl. Sci.*, vol. 13, no. 7, p. 4569, Apr. 2023, doi: 10.3390/app13074569.

[66] C. T. Cannon and S. Goericke, "Using Convolution Neural Networks to Develop Robust Combat Behaviors Through Reinforcement Learning," Naval Postgraduate School, Monterey, CA, 2020.

[67] J. R. Coble, "Optimal Naval Movement Simulation with Reinforcement Learning AI Agents," Naval Postgraduate School, Monterey, CA, 2023.

[68] M. G. Finley, "Applied Reinforcement Learning Wargaming With Parallelism, Cloud Integration, and AI Uncertainty," Naval Postgraduate School, Monterey, CA, 2023. [Online]. Available: https://calhoun.nps.edu/bitstream/handle/10945/72165/23Jun_Finley_Matthew.pdf?sequence=1&isAllowed=y

[69] J. T. Allen, "Enlisting AI in Course of Action Analysis as Applied to Naval Freedom of Navigation Operations," Naval Postgraduate School, Monterey, CA, 2022. [Online]. Available: https://hdl.handle.net/10945/71041

[70] K. Frans, J. Ho, X. Chen, P. Abbeel, and J. Schulman, "Meta Learning Shared Hierarchies." arXiv, Oct. 26, 2017. Accessed: Nov. 13, 2022. [Online]. Available: http://arxiv.org/abs/1710.09767

[71] C. Sammut and G. I. Webb, Eds., *Encyclopedia of Machine Learning*. New York ; London: Springer, 2010.

[72] S. J. Russell, P. Norvig, and E. Davis, *Artificial Intelligence: A Modern Approach*, 3rd ed. in Prentice Hall series in artificial intelligence. Upper Saddle River: Prentice Hall, 2010.

[73] G. Baldassarre and M. Mirolli, Eds., *Computational and Robotic Models of the Hierarchical Organization of Behavior*. Berlin, Heidelberg: Springer Berlin Heidelberg, 2013. doi: 10.1007/978-3-642-39875-9.

[74] R. S. Sutton, D. Precup, and S. Singh, "Between MDPs and semi-MDPs: A framework for temporal abstraction in reinforcement learning," *Artif. Intell.*, vol. 112, no. 1–2, pp. 181–211, Aug. 1999, doi: 10.1016/S0004-3702(99)00052-1.

[75] P. Dayan and G. E. Hinton, "Feudal Reinforcement Learning," p. 8, 1992.

[76] A. S. Vezhnevets *et al.*, "FeUdal Networks for Hierarchical Reinforcement Learning." arXiv, Mar. 06, 2017. Accessed: Sep. 28, 2022. [Online]. Available: http://arxiv.org/abs/1703.01161

[77] A. Levy, G. Konidaris, R. Platt, and K. Saenko, "Learning Multi-Level Hierarchies with Hindsight." arXiv, Sep. 03, 2019. Accessed: Sep. 09, 2022. [Online]. Available: http://arxiv.org/abs/1712.00948

[78] C. Florensa, Y. Duan, and P. Abbeel, "Stochastic Neural Networks for Hierarchical Reinforcement Learning." arXiv, Apr. 10, 2017. Accessed: Sep. 09, 2022. [Online]. Available: http://arxiv.org/abs/1704.03012

[79] A. P. Pope *et al.*, "Hierarchical Reinforcement Learning for Air-to-Air Combat." arXiv, Jun. 11, 2021. Accessed: Sep. 07, 2022. [Online]. Available: http://arxiv.org/abs/2105.00990

[80] Y. Wang, T. Jiang, and Y. Li, "A Hierarchical Reinforcement Learning Method on Multi UCAV Air Combat," in *2021 International Conference on Neural Networks, Information and Communication Engineering*, Z. Zhang, Ed., Qingdao, China: SPIE, Oct. 2021, p. 89. doi: 10.1117/12.2615268.

[81] Z. Zhang *et al.*, "Hierarchical Reinforcement Learning for Multi-agent MOBA Game." arXiv, Jun. 21, 2019. Accessed: Oct. 18, 2022. [Online]. Available: http://arxiv.org/abs/1901.08004

[82] S. Li, J. Zhang, J. Wang, Y. Yu, and C. Zhang, "Active Hierarchical Exploration with Stable Subgoal Representation Learning." arXiv, Mar. 05, 2022. Accessed: Oct. 18, 2022. [Online]. Available: http://arxiv.org/abs/2105.14750

[83] H. Wang, H. Tang, J. Hao, X. Hao, Y. Fu, and Y. Ma, "Large Scale Deep Reinforcement Learning in War-games," in *2020 IEEE International Conference on Bioinformatics and Biomedicine (BIBM)*, Seoul, Korea (South): IEEE, Dec. 2020, pp. 1693–1699. doi: 10.1109/BIBM49941.2020.9313387.

[84] C. Darken, "Atlatl," Monterey, CA, 2022. Available: https://gitlab.nps.edu/cjdarken/atlatl

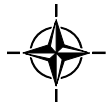